\documentclass{article}

\PassOptionsToPackage{numbers,compress}{natbib}
\usepackage[preprint]{neurips_2026}

\usepackage[utf8]{inputenc}
\usepackage[T1]{fontenc}
\usepackage{hyperref}
\usepackage{url}
\usepackage{booktabs}
\usepackage{amsmath}
\usepackage{amsfonts}
\usepackage{nicefrac}
\usepackage{microtype}
\usepackage{xcolor}
\usepackage{graphicx}
\usepackage{tabularx}
\usepackage{array}
\usepackage{bm}
\usepackage{float}
\usepackage{wrapfig}
\usepackage{enumitem}
\usepackage{algorithm}
\usepackage{algpseudocode}

\newcommand{\R}{\mathbb{R}}

\title{Efficient Factorized Volume Synthesis with \\ Lifted Inter-Slice Feature Trajectories}

\author{%
  Xinhe Zhang\textsuperscript{1,2} \\
  \texttt{xinhezhang@g.harvard.edu}
  \And
  Yuyang Zhang\textsuperscript{1,3} \\
  \texttt{yuyangzhang@g.harvard.edu}
  \AND
  Pengfei Jin\textsuperscript{2} \\
  \texttt{pjin1@mgh.harvard.edu}
  \And
  Arnau Marin-Llobet\textsuperscript{1} \\
  \texttt{amarinllobet@seas.harvard.edu}
  \And
  Na Li\textsuperscript{1,*} \\
  \texttt{nali@seas.harvard.edu}
  \And
  Quanzheng Li\textsuperscript{2,*} \\
  \texttt{li.quanzheng@mgh.harvard.edu}
}

\begin{document}

\maketitle
\begingroup
\renewcommand{\thefootnote}{\arabic{footnote}}
\footnotetext[1]{School of Engineering and Applied Sciences,
Harvard University.}
\footnotetext[2]{Center for Advanced Medical Computing and Analysis,
Massachusetts General Hospital and Harvard Medical School.}
\footnotetext[3]{Kempner Institute, Harvard University.}
\renewcommand{\thefootnote}{*}
\footnotetext[4]{Equal supervision.}
\renewcommand{\thefootnote}{}
\footnotetext[5]{Code is available at \url{https://github.com/xinhez/lift}.}
\endgroup

% Start subsequent numbered footnotes at 4.
\setcounter{footnote}{3}

\begin{abstract}
High-resolution 3D medical image generation remains challenging because fully volumetric models are computationally expensive, while efficient 2D slice generators often fail to preserve anatomical consistency across the third dimension. We propose LiFT, a framework for Lifted inter-slice Feature Trajectories that factorizes 3D volume synthesis into per-slice image generation and inter-slice trajectory learning. Rather than using a fully volumetric synthesis backbone, LiFT treats a volume as an ordered trajectory in feature space, capturing how anatomical structures appear, transform, and disappear across depth. We demonstrate the LiFT framework in both unconditional and conditional settings. In unconditional generation, a tri-planar drifting loss matches feature distributions between generated and real volumes across three orthogonal planes; in paired translation, a bidirectional \(z\)-context mixer trained with paired target supervision supplies through-plane coherence while preserving per-slice fidelity. We evaluate LiFT on BraTS 2023 for unconditional and missing-modality MR synthesis, and on SynthRAD2023 for MR-to-CT translation. LiFT achieves competitive image quality while retaining the efficiency of predominantly 2D synthesis, operating up to \(390\times\) faster than a volumetric diffusion baseline and consistently improving through-plane coherence through explicit inter-slice modeling. These results establish lightweight inter-slice trajectory learning as an efficient approach to high-resolution 3D medical synthesis.
\end{abstract}

\section{Introduction}

% Why 3D medical synthesis matters clinically
Medical image synthesis is fundamentally a three-dimensional (3D) problem because the anatomical and pathological targets of interest are volumetric. Tumors, ventricles, edema, white-matter tracts, skull boundaries, air cavities, and bone interfaces are not interpreted as isolated two-dimensional (2D) slices, but through their spatial continuity across a volume. This volumetric structure is also required by many downstream clinical and computational workflows, including segmentation, registration, dose planning, longitudinal monitoring, and image-guided intervention~\citep{chupetlovska_seg}. Generative models for 3D magnetic resonance (MR) and computed tomography (CT) imaging address data scarcity and data-sharing constraints~\citep{shin_gan_synth}, support augmentation and benchmarking~\citep{fridadar_gan, ibrahim_genai_review}, and enable cross-modality synthesis, such as missing-modality MR completion~\citep{acs_mri_review} or MR-to-CT synthesis for MR-only radiotherapy~\citep{roh_sct_review}. 

% Existing methods: fully 3D methods versus slice-wise or 2.5D methods

Existing generative approaches address this volumetric requirement in different ways. Fully 3D generators model inter-slice consistency directly, but often require substantial compromises in memory, resolution, architecture, or training strategy. Prior work has therefore explored hierarchical sub-volume training~\citep{hagan}, dense 3D denoising~\citep{ddpm3d, medddpm}, latent-space generation~\citep{brainldm, aldm}, wavelet-domain generation~\citep{Friedrich_2024, cwdm}, and patch-volume modeling~\citep{meddiff3d}. Despite their differences, these methods share the assumption that volumetric coherence should be learned primarily by a 3D generator. In contrast, slice-wise 2D synthesis can exploit mature high-resolution image-generation backbones and is often more efficient at preserving native in-plane detail. However, independently generated slices do not necessarily form anatomically coherent volumes. Recent two-and-a-half-dimensional~\citep{doubleunet}, adjacent-slice~\citep{bbdm}, multi-view~\citep{madm}, slice-based latent-diffusion~\citep{sliceldm}, and triplane-aware~\citep{tcamdiff} methods attempt to reduce this gap by injecting limited through-plane context into otherwise slice-oriented synthesis. This line of work highlights a central challenge: generating realistic slices is not sufficient unless they are also coordinated across depth.

% \Lina{the beginning sentence should revise. 1) One principle, every abbrevation should be fully spelled out when it firstly used in the main text. 2) better to do: Motivated by these observations, we propose ... 3) but it feels the motivation in the intro is too short with the first two paragraphs; on the other hand, this paragraph and the following paragraph is repetitive with ``Contribution". If we don't want to bring up our contribution too early, we should revise the paragraph in the way of ``motivating'' our main idea. For instance, remove the first sentence ``LiFT ...". Directly says, "In many medical" but please add more references so it reads in the way of motivating our main idea.} 

% The gap and motivation: strong 2D models can already synthesize good slices, but they lack through-plane organization
In many medical synthesis settings, high-resolution in-plane anatomy can be modeled effectively by 2D image-generation architectures. This is partly because clinical MR and CT volumes are often anisotropic, with substantially higher native resolution within each slice than across slices~\citep{saint, csam}, and partly because 2D backbones avoid the memory and optimization burden of fully volumetric generation~\citep{makeavolume, patchddm}. However, anatomical structures do not evolve independently from slice to slice. Tumors, edema, ventricles, cortical boundaries, bone interfaces, and air cavities follow continuous spatial trajectories through the volume. As a result, purely slice-wise synthesis can preserve local image detail yet produce through-plane discontinuities and inconsistent anatomy~\citep{doubleunet, bbdm}. These observations motivate a factorized view of medical volume synthesis, in which high-resolution slice formation and through-plane anatomical organization are treated as related but separable problems.

% Introducing new method
\begin{figure}[t]
    \centering
    \includegraphics[width=\textwidth]{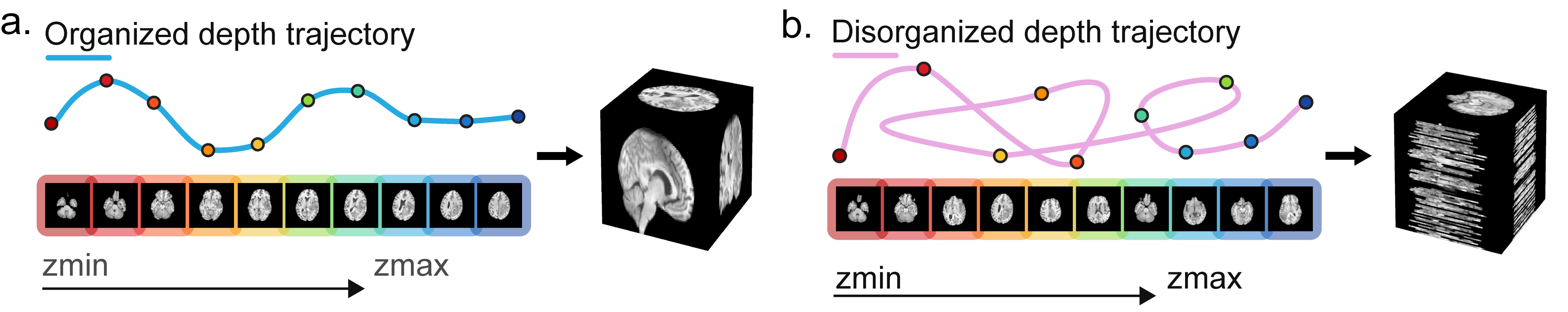}
    \caption{Conceptual motivation: LiFT organizes slice representations into coherent depth trajectories.}
    \label{fig:overview}
\end{figure}

Motivated by these observations, we propose \textbf{LiFT}, a framework for Lifted inter-slice Feature Trajectories in medical volume synthesis. As illustrated in Fig.~\ref{fig:overview}, we view a volume as an ordered trajectory through slice-level feature space: plausible individual slices must also evolve coherently with depth to form a valid 3D volume. LiFT therefore keeps the core synthesis pathway predominantly 2D while learning a lightweight mechanism that organizes slice representations along the through-plane axis. The form of this through-plane organization depends on the supervision available. In the unconditional setting, where no paired target volume exists for each generated sample, LiFT encourages coherence distributionally through tri-planar drifting. In paired translation, where registered target volumes are available, LiFT learns depth-axis context directly under supervised reconstruction objectives. 

% \Lina{better to just have one bold phrase ``contribution''. Don't do `` cope of contribution'' and ``contributions''}
% \Lina{the contribution bullets: 1) the first one should be more detailed; 2) the other three points are all ``emperical studies''. I would oraganize them in one big bullet point and then put a flow in organize these three emperical studies.}

\paragraph{Contributions.}
We propose LiFT, a 2D-to-volume synthesis framework for medical image generation and translation. Our main contributions are:

\begin{itemize}[leftmargin=*]
    \item \textbf{Factorized 2D-to-volume synthesis.} We formulate volumetric synthesis as high-capacity 2D generation plus lightweight inter-slice trajectory modeling, with supervision adapted to unconditional and paired synthesis settings.
    \item \textbf{Empirical validation across complementary clinical synthesis settings.} We evaluate the same lifting principle across unconditional MR generation, missing-modality MR synthesis, and MR-to-CT synthesis. The experiments quantify image quality, volumetric coherence, parameter count, and inference cost relative to volumetric and slice-wise baselines, demonstrating that lightweight through-plane modeling preserves the computational advantages of predominantly 2D synthesis across distinct supervision regimes.
\end{itemize}

\section{Related Work}

% \Lina{this paragraph didn't conclude remaining problems and challenges or connect back to ``LiFT". }
\paragraph{Volumetric and slice-wise medical synthesis.}
Volumetric methods enforce 3D structure by operating on 3D tensors or compressed 3D representations, including hierarchical sub-volume generative adversarial networks (GANs)~\citep{hagan}, 3D and conditional denoising diffusion probabilistic models (DDPMs)~\citep{ddpm3d, medddpm}, latent generative models~\citep{brainldm, aldm}, wavelet-domain models~\citep{Friedrich_2024, cwdm}, and patch-volume latent diffusion models (LDMs)~\citep{meddiff3d}. These approaches directly model inter-slice dependence, but their volumetric design often requires compromises in memory, spatial resolution, training strategy, or architectural complexity~\citep{patchddm}. Slice-wise alternatives instead synthesize volumes by stacking independent 2D outputs from pix2pix-style or CycleGAN-style backbones~\citep{pix2pix, cyclegan, maskgan, doubleunet}, with recent extensions adding adjacent-slice alignment, multi-view averaging, or slice-aggregation modules to recover coherence~\citep{bbdm, madm, sliceldm}. These methods preserve the efficiency and native in-plane resolution of 2D synthesis, but the resulting volumes can still suffer from discontinuities or inconsistent anatomy across depth~\citep{doubleunet, bbdm}. Synthetic CT is a representative application of this tradeoff: SynthRAD2023 benchmarks MR-to-CT and cone-beam-CT-to-CT synthesis under image- and dose-based metrics~\citep{synthrad2023}, and patch-based conditional GANs, CycleGANs, U-Nets, and drifting-model variants have all been applied to synthetic-CT generation~\citep{multiplanar_cgan, cycle_sct, farjam_sct, mri_ct_drifting}. Across these settings, the remaining challenge is to obtain volumetric coherence without paying the full cost of a fully volumetric generator or relying on independent slice synthesis followed by post hoc consistency correction.

\paragraph{Factorized 2D-to-3D and feature-distribution methods.}
Closer to LiFT are methods that reduce the cost of volumetric generation by combining 2D and 3D components. Make-A-Volume introduces volumetric layers into a 2D latent diffusion backbone~\citep{makeavolume}, EG3D decouples feature generation from neural rendering through structured three-dimensional-aware representations~\citep{eg3d}, and TCAM-Diff uses triplane-aware cross-attention diffusion for medical volumes~\citep{tcamdiff}. These methods demonstrate that dense 3D generation is not the only way to obtain volumetric structure. However, they still build 3D awareness into the main generative pathway, either by modifying the backbone, introducing specialized volumetric representations, or coupling synthesis tightly to a 3D rendering or attention mechanism. This leaves a different question open: when a strong 2D synthesizer already produces high-quality in-plane anatomy, can volumetric coherence be added without redesigning the generator itself?

LiFT addresses this question by treating 3D synthesis as a lifting problem rather than a fully 3D generation problem. The main synthesizer remains predominantly 2D and retains most of the representational capacity, while a small depth-indexed feature trajectory module supplies the missing through-plane organization. On the supervisory side, LiFT-U is related to perceptual losses~\citep{johnson_perceptual}, LPIPS~\citep{lpips}, feature matching~\citep{salimans_fm}, moment-matching kernels~\citep{gmmn, mmdgan}, and drifting losses~\citep{drifting, latentdrift}, which align samples in learned feature spaces rather than requiring pointwise correspondence. Unlike these generic feature-space objectives, LiFT-U applies distributional matching to tri-planar anatomical trajectories of completed volumes, making the supervision explicitly volume-structured. In the paired setting, LiFT-C uses the same lifting principle but replaces distributional trajectory matching with direct supervised depth-axis context modeling. Thus, LiFT differs from prior factorized methods not by introducing another heavy 3D generator, but by isolating the minimum learned through-plane mechanism needed to organize a strong 2D synthesizer into a coherent volume.

\section{Problem Formulation}
\label{sec:problem}

We consider volumetric medical image synthesis settings in which a 3D volume is represented as a stack of 2D slices. This formulation covers both unconditional volume generation and paired image-to-image translation tasks, including missing-modality MRI synthesis and MR-to-CT synthesis.

Let a volume be denoted by
$V \in \R^{C \times H \times W \times D}$,
% \Lina{should explain what $C, H, W$ stand for. Especially $C$ is important to explain}
where $C$ is the number of image channels or modalities, $H$ and $W$ are the in-plane spatial dimensions, and $D$ indexes the through-plane axis along which 2D slices are stacked. We denote the slice at depth $d$ by
$v_d \in \R^{C \times H \times W}$.
We study two volumetric synthesis regimes that share this representation but differ in the supervision available to the model.

\paragraph{Unconditional 3D synthesis.}
Given a target distribution $p_{\mathrm{data}}(V)$ over real volumes, the goal is to generate samples
$\hat{V} \sim p_{\mathrm{data}}$.
In this setting, each generated volume has no paired ground-truth counterpart. Therefore, whole-volume realism cannot be supervised by a direct voxel-wise comparison to a target volume; instead, it must be encouraged through distributional objectives that compare generated volumes with real volumes.
% \Lina{``so volume-level ... distributional", is confusing to readers: not sure why this sentence pops up and what is referring to.}

\paragraph{Paired 3D translation.}
% \Lina{add transition words. You might bring missing-modality MRI and MR-to-CT at the beginning to motivate.}
In contrast, many clinical synthesis tasks provide registered source--target pairs. Given a pair
$(X, Y) \in
\R^{C_{\mathrm{in}}\times H\times W\times D}
\times
\R^{C_{\mathrm{out}}\times H\times W\times D}$,
the goal is to predict $\hat{Y} \approx Y$ from $X$.
% \Lina{need to explain $C_{\text{in}}$ and $C_{\text{out}}$}
Here, $C_{\mathrm{in}}$ denotes the number of input channels or source modalities, such as different contrasts of MR, and $C_{\mathrm{out}}$ denotes the number of output channels or target modalities. Both missing-modality MRI synthesis, where the model predicts an unobserved MRI modality from available MRI modalities, and MR-to-CT synthesis, where the model predicts CT from MR, fit this regime. Because the target volume $Y$ is available, direct voxel-level and feature-level supervision can be applied.

In both regimes, a synthesized volume should satisfy two requirements: (i) per-slice anatomical realism, and (ii) through-plane coherence under coronal and sagittal reformats. Our goal is to satisfy both requirements without incurring the memory and compute cost of a fully volumetric generator.

\section{Method}
\label{sec:method}

\begin{figure}[!ht]
  \centering
  \includegraphics[width=\textwidth]{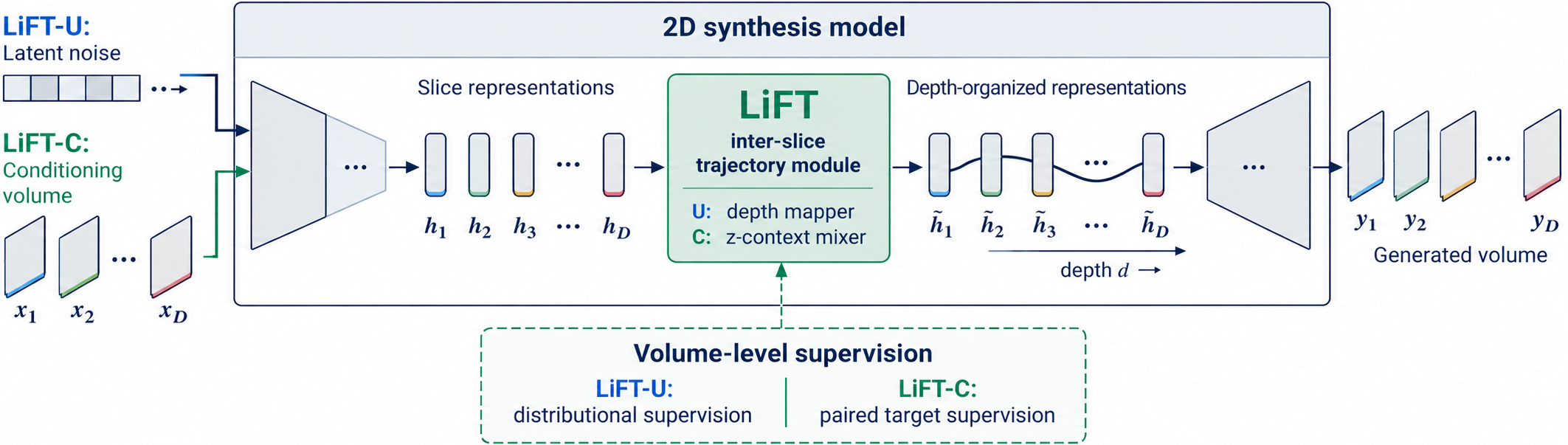}
  \caption{
    % \Lina{this figure resolution is not good. Should use PDF format (vector format) to save the figure; Moreoever, words in the figure should be larger. It is difficult to read. }
    LiFT factorizes volume synthesis into a 2D synthesizer and an inter-slice trajectory module.}
  \label{fig:lift_pipeline}
\end{figure}

% \Lina{the beginning of the section looks strange. Section 3 didn't provide any methods it. The beginning of the section should to warm up the introduction of LiFT-U. }
LiFT converts a strong 2D synthesizer into a volume generator by separating in-plane synthesis from through-plane organization. As summarized in Fig.~\ref{fig:lift_pipeline}, the main synthesis pathway remains two-dimensional, while a lightweight inter-slice trajectory module organizes slice-level representations across depth. The exact form of this module depends on the synthesis regime: LiFT-U uses a depth mapper to generate a sequence of depth-dependent conditioning codes from latent noise, whereas LiFT-C uses a bidirectional $z$-context mixer to exchange information among the bottleneck representations of a conditioning volume. The resulting depth-organized representations are synthesized slice-wise and assembled into the generated volume.

The supervision likewise adapts to the available data. LiFT-U learns the trajectory module distributionally from unpaired real volumes, while LiFT-C uses paired target supervision from registered source--target volumes. The following sections describe the two realizations in detail.

\subsection{LiFT-U: distributional lifting for unconditional synthesis}
\label{sec:method:liftu}

For unconditional synthesis, LiFT-U couples a frozen 2D slice generator
with a learned depth mapper and trains the resulting volume using
distributional trajectory supervision.

\paragraph{Frozen 2D slice generator.}
We first pretrain a 2D generator $G_{2\mathrm{D}}(\cdot;\theta)$ to synthesize
axial brain MR slices, then freeze $\theta$. The frozen generator maps a
per-slice conditioning vector $c_d \in \R^{d_c}$ to a 2D slice,
\begin{equation}
    \hat{v}_d = G_{2\mathrm{D}}(c_d;\theta).
\end{equation}
Freezing the slice generator preserves the in-plane anatomical prior and
confines all subsequent training to through-plane organization.

\paragraph{Depth mapper.}
The trajectory module $M_\phi$ is a lightweight network that, given a global
latent code $\xi\sim\mathcal N(0,I)$ and a Fourier
encoding~\citep{fourier_features} $\gamma(d)$ of the depth coordinate,
returns a per-slice conditioning vector,
\begin{equation}
    c_d = M_\phi(\xi,\gamma(d)).
\end{equation}
The generated volume is
\begin{equation}
    \hat{V}(\xi;\phi)
    =
    \operatorname{Stack}_{d=1}^{D}
    G_{2\mathrm{D}}(c_d;\theta).
    \label{eq:liftu_volume}
\end{equation}

\paragraph{Tri-planar drifting loss.}
Let
$\Pi=\{\pi_{\mathrm{ax}},\pi_{\mathrm{cor}},\pi_{\mathrm{sag}}\}$
denote the axial, coronal, and sagittal slicing operators. For each plane
$\pi\in\Pi$, let $\mathcal{S}_{\pi}$ denote its valid slice-index set, and let
$E_{\mathrm{feat}}$ be a fixed feature extractor that maps a 2D slice to a
feature vector in $\R^{d_f}$. For a generated volume $\hat{V}$ and an
empirical pool of real volumes $\mathcal{D}_{\mathrm{real}}$, define the real
feature bank for plane $\pi$ and slice index $s$ as
\begin{equation}
    \mathcal{B}_{\pi,s}
    \;=\;
    \bigl\{
    E_{\mathrm{feat}}(\pi_s(V_j))
    \;:\;
    V_j \in \mathcal{D}_{\mathrm{real}}
    \bigr\}.
    \label{eq:feature_bank}
\end{equation}

Given a generated slice feature
$g = E_{\mathrm{feat}}(\pi_s(\hat{V}))$, let
$\mathcal{G}_{\pi,s}$ contain the features of the other generated volumes
in the minibatch at the same plane and slice index. The drifting
operator~\citep{drifting} attracts $g$ toward real features and repels it
from other generated features using kernel-weighted local means:
\begin{equation}
    \begin{aligned}
    \mu(g;\mathcal{X})
    &=
    \frac{
        \sum_{r \in \mathcal{X}}
        K\bigl(\|g-r\|/R\bigr)\,r
    }{
        \sum_{r \in \mathcal{X}}
        K\bigl(\|g-r\|/R\bigr)
    }, \\[4pt]
    \operatorname{Drift}
    \bigl(g,\mathcal{B}_{\pi,s},\mathcal{G}_{\pi,s}\bigr)
    &=
    \bigl\|
        g-\operatorname{sg}\bigl(
        g+\mu(g;\mathcal{B}_{\pi,s})
        -\mu(g;\mathcal{G}_{\pi,s})
        \bigr)
    \bigr\|_2^2,
    \end{aligned}
    \label{eq:drift_op}
\end{equation}
where $K$ is a Laplacian kernel, $R$ is its annealed bandwidth, and
$\operatorname{sg}(\cdot)$ denotes stop-gradient. The LiFT-U objective
averages this drift over generated volumes, projection planes,
and slice indices:
\begin{equation}
    \mathcal{L}_{\text{LiFT-U}}(\phi)
    \;=\;
    \frac{1}{|\Pi|}
    \sum_{\pi \in \Pi}
    \mathbb{E}_{
        \xi \sim \mathcal{N}(0,I),\,
        s \sim \mathrm{Unif}(\mathcal{S}_{\pi})
    }
    \Bigl[
    \operatorname{Drift}\bigl(
        E_{\mathrm{feat}}(\pi_s(\hat{V}(\xi;\phi))),\,
        \mathcal{B}_{\pi,s},\,
        \mathcal{G}_{\pi,s}
    \bigr)
    \Bigr].
    \label{eq:liftu_loss}
\end{equation}

\subsection{LiFT-C: supervised lifting for paired translation}
\label{sec:method:liftc}
% \Lina{check my suggestion for section 4.1. Right now, it is difficult to understand both section 4.1 and 4.2}

For paired translation, LiFT-C adds inter-slice context at the spatial
bottleneck of an otherwise 2D encoder--decoder, allowing each predicted
slice to incorporate full-volume context while leaving the encoder, decoder,
and skip pathways two-dimensional.

\paragraph{Conditional 2D translator.}
Let $G_\theta = D_\theta \circ E_\theta$ denote a 2D encoder--decoder applied to source slices $x_d = X_d$. The encoder produces a spatial bottleneck feature map
\begin{equation}
    h_d = E_\theta(x_d),
\end{equation}
which retains the within-slice spatial organization of the input. In missing-MR synthesis, $x_d$ contains the available MR contrasts and $Y$ is the missing contrast. In MR-to-CT synthesis, $x_d$ is the MR input and $Y$ is the registered CT target.
\paragraph{Spatial bidirectional $z$-context mixer.}
A slice-wise translator processes each bottleneck feature independently and
therefore cannot exchange information across depth. LiFT-C applies a
bidirectional recurrent context module $M_\phi^C$ to the sequence of spatial
bottleneck features:
\begin{equation}
    (c_1,\ldots,c_D)
    =
    M_\phi^C\!\left(h_{1:D}\right),
    \label{eq:liftc_context}
\end{equation}
where the recurrent module propagates information along the slice dimension
while preserving the spatial layout of the bottleneck representation,
producing a spatial context map $c_d$ for each slice.

When memory constraints require processing only a subset of the $z$-stack
during optimization, we additionally provide an absolute positional encoding
$p(d)$ to identify each slice's location within the full volume.

The context map is merged with the corresponding 2D bottleneck through a residual connection,
\begin{equation}
    \tilde h_d = h_d + R_\phi(c_d),
    \qquad
    \hat y_d = D_\theta(\tilde h_d),
    \qquad
    \hat Y = \operatorname{Stack}_{d=1}^{D}\hat y_d .
    \label{eq:liftc_decode}
\end{equation}

Thus, all cross-slice computation is confined to the bottleneck; the encoder,
decoder, and skip pathways remain two-dimensional.

\paragraph{Paired volumetric supervision.}
Because registered target volumes are available, LiFT-C is trained with
direct volumetric supervision:
\begin{equation}
    \mathcal{L}_{\mathrm{LiFT\text{-}C}}
    =
    \mathcal{L}_{\mathrm{sup}}(\hat Y,Y),
    \label{eq:liftc_loss}
\end{equation}
where $\mathcal{L}_{\mathrm{sup}}$ denotes the task-specific supervised
objective defined using the predicted and registered ground-truth volumes.
This supervision jointly constrains reconstruction and variation across the
depth axis. Task-specific training details are summarized in
Appendix~\ref{sec:appendix:hyperparams}.

\section{Experiments}
\label{sec:exp}

We evaluate LiFT on three 3D medical synthesis tasks that test whether a 2D synthesizer with a lightweight trajectory or context module can recover useful volumetric structure. Tasks~A and~B use BraTS~2023 GLI~\citep{medperf, brats2021, brats_menze, brats_bakas, tcga_gbm, tcga_lgg} for unconditional 3D brain MR generation and missing-modality MR synthesis, respectively. Task~C uses SynthRAD2023~\citep{synthrad2023} for MR-to-CT synthesis.

All models trained or reproduced by us are run on a single NVIDIA RTX~5090 GPU,
and all locally measured inference times are obtained on this hardware. Per-task hyperparameters, preprocessing details, and dataset/model licenses are provided in Appendices~\ref{sec:appendix:hyperparams} and~\ref{sec:appendix:licenses}.

\subsection{Task A: unconditional brain MR generation}
\label{sec:exp:uncond}

Task~A evaluates LiFT-U on unconditional 3D brain MR synthesis using the BraTS 2023 GLI cohort. The model generates T1n volumes at $128\times128\times128$ resolution. Following the WDM evaluation setting~\citep{Friedrich_2024}, volumes are skull-stripped, intensity-normalized to $[-1,1]$, and resampled to an isotropic $128^3$ grid.

We instantiate LiFT-U as described in Section~\ref{sec:method}: an axial 2D slice generator is trained first and then frozen while a lightweight depth mapper organizes generated slices into coherent volumes. For this task, the depth mapper has 1.455M trainable parameters, and the tri-planar drift loss uses features from a frozen ImageNet-pretrained ResNet-18. Optimizer, learning rate, and batch construction are summarized in Appendix~\ref{sec:appendix:hyperparams}.

We evaluate unconditional generation using distributional image quality and generation efficiency. Following the WDM evaluation protocol~\citep{Friedrich_2024}, we compute Fr\'echet Inception Distance (FID)~\citep{fid_heusel} from 2048-dimensional MedicalNet ResNet-50 features~\citep{medicalnet}. For LiFT-U, MAISI-v2~\citep{maisiv2}, and 3D MedDiffusion~\citep{meddiff3d}, FID is computed from $1{,}000$ generated volumes against the same $1{,}000$-volume real reference set. We additionally report inference-time parameter count and wall-clock inference time per $128^3$ volume. For WDM, we report their published FID and measure efficiency from the released model under the same hardware setting. Evaluating the released model under our protocol gives $0.178$.

\begin{table}[H]
\centering
\caption{Unconditional T1n generation on BraTS~2023 GLI.}
\label{tab:unconditional_quality_efficiency}
\begin{tabular}{lccc}
\toprule
Method & FID $\times 10^3$ $\downarrow$ & Params. (M) $\downarrow$ & Time / vol. (s) $\downarrow$ \\
\midrule
WDM & 0.154 & 57.9 & 26.92 \\
MAISI-v2 & 0.294 & 201.4 & 0.506 \\
3D MedDiffusion & \textbf{0.043} & 166.3 & 15.37 \\
LiFT-U & 0.066 & \textbf{30.3} & \textbf{0.039} \\
\bottomrule
\end{tabular}
\end{table}

\begin{figure}[!ht]
\centering
\includegraphics[width=\textwidth]{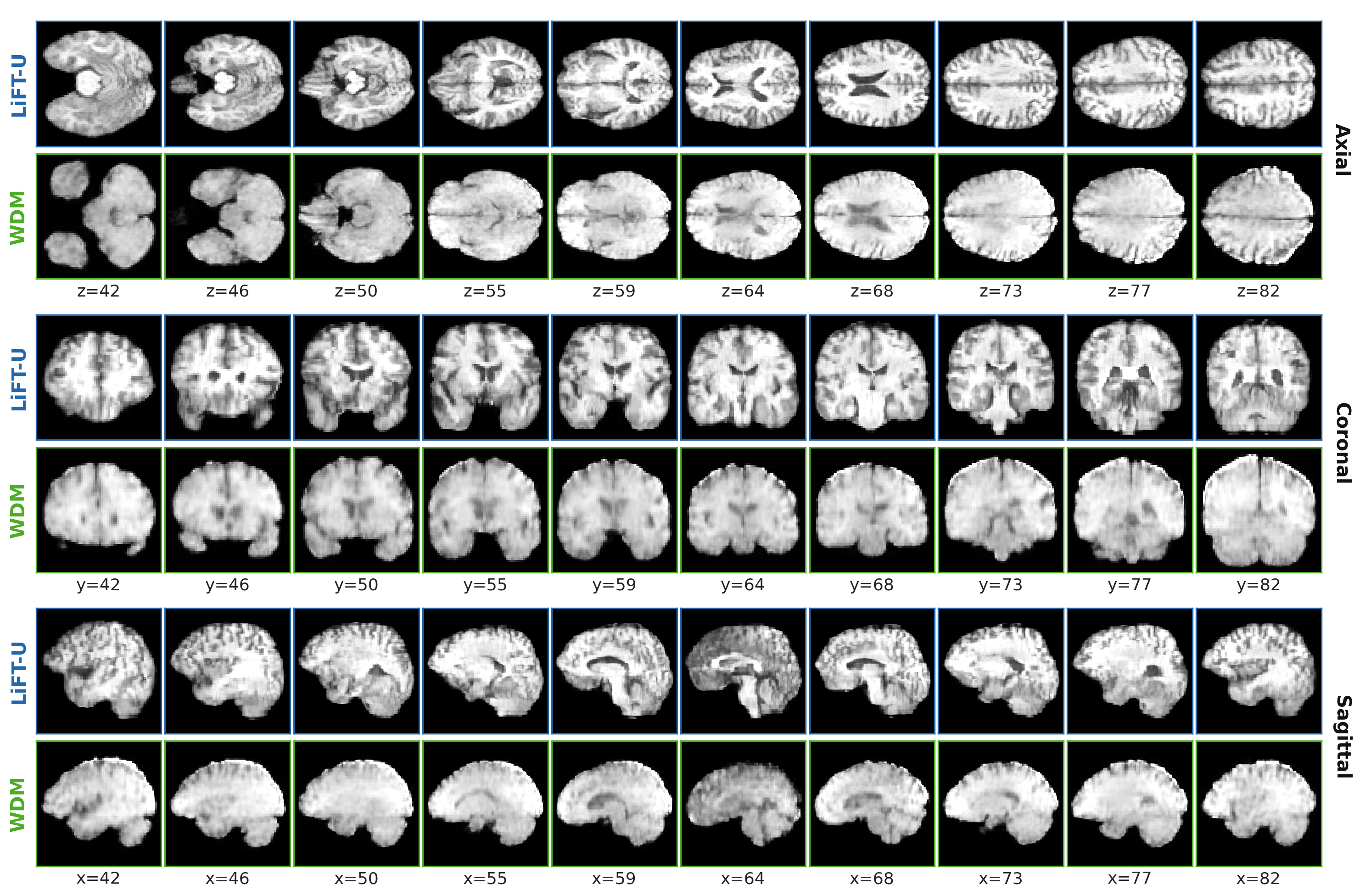}
\caption{Qualitative unconditional MR comparison between LiFT-U and WDM.}
\label{fig:qual_liftu_main}
\end{figure}

Table~\ref{tab:unconditional_quality_efficiency} shows the quality--efficiency trade-off. Relative to 3D MedDiffusion, LiFT-U is approximately $390\times$ faster and uses $5.5\times$ fewer parameters. Relative to MAISI-v2, LiFT-U achieves lower FID with $6.6\times$ fewer parameters and $13\times$ lower inference time. Figure~\ref{fig:qual_liftu_main} shows representative axial, coronal, and sagittal reformats from LiFT-U and WDM across depth, providing a qualitative view of the generated volumetric anatomy.

\subsection{Task B: missing-modality MR synthesis}
\label{sec:exp:cond_mr}

Task~B evaluates LiFT-C on missing-modality MR synthesis using the BraTS 2023 GLI cohort under the cWDM protocol~\citep{cwdm}. Given any three of the four contrasts T1n, T1c, T2w, and T2f, the model synthesizes the fourth at native $240\times240\times155$ resolution. Volumes are clipped to the $0.1$--$99.9$ percentile and normalized to $[0, 1]$; predictions are clipped to $[0, 1]$ and zeroed outside the brain mask (where the target is zero).

We instantiate LiFT-C as described in Section~\ref{sec:method:liftc}.
The conditional translator is a 2D U-Net with the three available source
modalities channel-concatenated as input. Its spatial context module operates
on the $512\times15\times15$ bottleneck feature maps from all $155$ slices.
A two-layer bidirectional GRU propagates information along depth independently
at each bottleneck location, using sinusoidal absolute-position encodings.
The resulting context is injected through a zero-initialized residual
$1\times1$ projection. The model is trained with direct supervision from the
registered target volume; additional training details are given
in Appendix~\ref{sec:appendix:hyperparams}.

Evaluation uses the BraTS~2023 GLI $219$-subject validation split.
LiFT-C, pix2pix, cWDM, and Fast-cWDM are evaluated on the same subjects using
identical preprocessing and metric implementations. We report average
reconstruction quality across the four target contrasts together with
inference-time parameter count and wall-clock inference time. cWDM uses
$1000$ diffusion steps and Fast-cWDM uses $100$, while LiFT-C performs
deterministic non-iterative synthesis.

The purely 2D pix2pix baseline is fastest (Table~\ref{tab:missing_mr}),
but LiFT-C improves average PSNR by $2.066$\,dB and SSIM by $0.0183$.
LiFT-C differs from full cWDM by only $0.008$\,dB in average PSNR while
requiring approximately $133\times$ less inference time, and is
$13.3\times$ faster than Fast-cWDM while using fewer inference-time
parameters. Per-contrast PSNR and SSIM results are reported in Appendix~\ref{sec:appendix:missing_mr_percontrast}.

\begin{table}[!ht]
\centering
\caption{Missing-MR synthesis on BraTS~2023 GLI.}
\label{tab:missing_mr}
\begin{tabular}{lcccc}
\toprule
Method & Avg. PSNR $\uparrow$ & Avg. SSIM $\uparrow$
       & Params. (M) $\downarrow$ & Time / vol. (s) $\downarrow$ \\
\midrule
pix2pix   & 26.351 & 0.9345 & 54.4 & \textbf{0.053} \\
Fast-cWDM & 28.233 & 0.9456 & 81.5 & 15.54 \\
cWDM      & \textbf{28.425} & 0.9525 & 81.5 & 154.8 \\
LiFT-C    & 28.417 & \textbf{0.9528} & \textbf{41.3} & 1.166 \\
\bottomrule
\end{tabular}
\end{table}
 
We compare the same trained LiFT-C checkpoint with the context mixer
enabled or disabled at inference. Enabling context reduces all three
through-plane inconsistency measures, including a $48.4\%$ mean relative
reduction in the axial-bias SSIM gap across contrasts. Full results and
paired statistics are provided in
Appendix~\ref{sec:appendix:taskb_context}.

\subsection{Task C: MR-to-CT synthesis}
\label{sec:exp:mrct}

Task~C evaluates LiFT-C on paired MR-to-CT synthesis using SynthRAD2023 Task~1 Brain~\citep{synthrad2023}. We use a fixed patient-level $80/20$ split of the $180$ subjects: $144$ subjects for training and $36$ subjects for model selection and evaluation. The same split is used for all compared models. The input is a T1-weighted MR volume and the target is the registered CT. Following the preprocessing pipeline used for the CBAM3D-UNet baseline~\citep{cbam3dunet}, MR volumes are z-score-normalized over the full volume, CT volumes are clipped to $[-1000, 2000]$\,HU and scaled to $[-1, 1]$, and both modalities are cropped to $128^3$ using the SynthRAD brain-mask bounding-box center.

We instantiate LiFT-C as a residual variant of
Section~\ref{sec:method:liftc}. A frozen axial 2D translator first predicts
a baseline CT. LiFT-C then operates on its spatial bottleneck representation
using a single-layer bidirectional GRU with hidden size $128$, applied
independently across spatial locations along the depth axis. Fourier
positional encoding with eight frequencies is concatenated to the recurrent
input at each depth. The resulting context is injected through a
zero-initialized convolution with a learnable gate, and the
context module predicts an additive residual correction to the 2D translator
output.

We evaluate all methods on the same $36$-subject model-selection/evaluation
cohort using the official SynthRAD masked evaluator for MAE, PSNR, SSIM,
and NCC, reporting subject-level results as mean$\pm$SD. Through-plane consistency is evaluated separately using
adjacent-slice $\Delta_z$ MAE and correlation. Bone- and air-region $\Delta_z$ errors separately measure consistency across high-contrast anatomical structures. Region-mask definitions are provided in Appendix~\ref{sec:appendix:metrics}. Baselines, run by us on the same split, are a 3D CBAM-attention U-Net (CBAM3D-UNet)~\citep{cbam3dunet}, reimplemented from the published architecture description, and axial 2D pix2pix-U-Net and pix2pix-ResNet.

On conventional image-quality metrics, LiFT-C obtains the best mean value
among the evaluated methods on all four metrics (Table~\ref{tab:taskc_image_metrics}).
Its performance is close to the strongest 2D baseline, pix2pix-ResNet:
the differences in MAE, PSNR, and NCC are not significant after Bonferroni
correction, while the SSIM improvement remains significant. Relative to the
context-disabled 2D translator, LiFT-C improves all four metrics
consistently across subjects.

\begin{table}[!ht]
\centering
\small
\caption{Conventional MR-to-CT image metrics.}
\label{tab:taskc_image_metrics}
\begin{tabular}{lcccc}
\toprule
Method & MAE (HU) $\downarrow$ & PSNR $\uparrow$ & SSIM $\uparrow$ & NCC $\uparrow$ \\
\midrule
Pix2pix-UNet
& $74.062\pm11.136$
& $26.934\pm1.165$
& $0.8872\pm0.0241$
& $0.8119\pm0.0580$ \\
Pix2pix-ResNet
& $55.207\pm8.914$
& $29.080\pm1.222$
& $0.9287\pm0.0166$
& $0.8829\pm0.0376$ \\
CBAM3D-UNet
& $60.938\pm9.321$
& $28.297\pm1.303$
& $0.9201\pm0.0170$
& $0.8642\pm0.0483$ \\
\midrule
Context disabled
& $56.007\pm9.337$
& $28.842\pm1.434$
& $0.9301\pm0.0164$
& $0.8764\pm0.0460$ \\
LiFT-C
& $\mathbf{54.093\pm9.323}$
& $\mathbf{29.096\pm1.510}$
& $\mathbf{0.9355\pm0.0158}$
& $\mathbf{0.8842\pm0.0446}$ \\
\bottomrule
\end{tabular}
\end{table}

For through-plane evaluation, $\Delta_z$ denotes the forward adjacent-slice
difference along depth; lower $\Delta_z$ MAE and higher $\Delta_z$
correlation indicate better agreement with the reference volume. The
$\Delta_z$ metrics are computed with our adjacent-slice coherence evaluator,
separately from the official SynthRAD image-quality evaluator. The effect of
through-plane context is substantially larger on these depth-sensitive
metrics. Relative to the context-disabled model, full-volume, bone, and air $\Delta_z$ MAE
decrease by $10.2\%$, $11.1\%$, and $13.1\%$, respectively, while
$\Delta_z$ correlation increases by $7.9\%$ (Table~\ref{tab:mrct_coherence}).
The improvements are consistent at the subject level; paired statistics are
reported in Appendix~\ref{sec:appendix:mrct_stats}.

\begin{table}[!ht]
\centering
\caption{Through-plane coherence for MR-to-CT synthesis.}
\label{tab:mrct_coherence}
\setlength{\tabcolsep}{4pt}
\begin{tabular}{lcccc}
\toprule
Method & $\Delta_z$ MAE Full $\downarrow$ & $\Delta_z$ MAE Bone $\downarrow$ & $\Delta_z$ MAE Air $\downarrow$ & $\Delta_z$ corr $\uparrow$ \\
\midrule
Pix2pix-UNet & 52.20 & 180.23 & 199.71 & 0.575 \\
Pix2pix-ResNet & 44.91 & 159.54 & 186.91 & 0.650 \\
CBAM3D-UNet & 39.88 & 141.10 & \textbf{142.37} & 0.713 \\
\midrule
Context disabled & 42.73 & 151.73 & 172.21 & 0.675 \\
LiFT-C & \textbf{38.369} & \textbf{134.968} & 149.722 & \textbf{0.729} \\
\bottomrule
\end{tabular}
\end{table}

\section{Discussion and Conclusion}
\label{sec:limitations}

Our results suggest that volumetric synthesis can
benefit from separating in-plane image formation from through-plane
organization. Across the evaluated tasks, LiFT retains a predominantly
2D synthesis pathway while learning inter-slice dependencies through
a lightweight trajectory module. LiFT-U achieves a favorable
quality--efficiency trade-off against volumetric generators, and
LiFT-C approaches diffusion-based missing-MR reconstruction quality
at substantially lower inference cost. These findings support
factorization as a practical design principle for medical volume
synthesis.

The paired experiments clarify the role of inter-slice modeling: enabling context reduces all three through-plane inconsistency measures in missing-MR synthesis, while yielding larger relative improvements on depth-sensitive than conventional image-quality metrics in MR-to-CT synthesis. This pattern suggests that context primarily coordinates anatomy across slices, a benefit less apparent in aggregate reconstruction scores. Evaluating both in-plane fidelity and through-plane consistency is therefore important for understanding volumetric context.

The unconditional and paired settings also illustrate two ways to
learn this organization under different supervision. LiFT-U uses
tri-planar distributional matching to constrain generated volumes
without requiring paired targets. LiFT-C instead learns depth-axis
context from registered source--target volumes. Their complementary
results indicate that the factorization principle can accommodate
different learning objectives, with the form of inter-slice modeling
adapted to the available supervision.

This separation also defines the scope of LiFT. Its performance
remains coupled to the underlying 2D backbone: inter-slice coordination
can improve volumetric consistency, while anatomical detail, contrast
recovery, and bone/air interfaces remain dependent on the synthesis
model. Residual errors in bone and air regions in MR-to-CT synthesis
illustrate this dependence. LiFT-U's distributional evaluation
characterizes generated sample quality without establishing
anatomical correctness for individual volumes, while LiFT-C depends
on the quality of paired registration and preprocessing.

Our experiments focus on brain imaging, with missing-MR synthesis
evaluated on a shared $219$-subject validation cohort. Extending
this approach to deformable anatomy and variation across institutions,
scanners, acquisition protocols, and pathology distributions will
help establish its broader applicability. For MR-to-CT synthesis,
further evaluation of dose accuracy, contouring, registration, and
reader assessment would connect the observed image-domain
improvements to clinical utility. 

LiFT demonstrates the value of learning through-plane organization as a distinct component of medical image synthesis. By combining the capacity of 2D backbones with lightweight inter-slice modeling, it offers an efficient route to high-resolution volume generation across unconditional and paired settings. The paired experiments further show that the contribution of inter-slice context is more clearly revealed by evaluating volumetric coherence alongside conventional image-quality metrics. Together, these results support factorizing in-plane synthesis and through-plane organization as a practical approach to balancing image quality, volumetric coherence, and computational efficiency.

\paragraph{Broader impact and safety.}
Medical image synthesis can support research and data augmentation,
while carrying risks related to dataset bias, alteration of
acquisition-specific features, fabrication of clinical evidence,
and patient privacy. Our experiments evaluate research benchmark
outputs; clinical deployment requires further validation.
Nearest-neighbor retrieval for unconditional generation shows no
evidence of near-exact training-volume copying among the evaluated samples
(Appendix~\ref{sec:appendix:memorization}).
 
\newpage

\section*{Acknowledgements}
This work was supported by the National Institutes of Health (NIH) under award number R01HL159183.
Yuyang Zhang is supported by the Kempner Institute Graduate Research Fellowship.
Arnau Marin-Llobet is supported by Coefficient Giving and the RCC-Harvard Fellowship.

\bibliographystyle{unsrtnat}
\bibliography{references}

\newpage
\appendix

\section{Qualitative comparisons}
\label{sec:appendix:qualitative}

\begin{figure}[!ht]
\centering
\includegraphics[width=\textwidth]{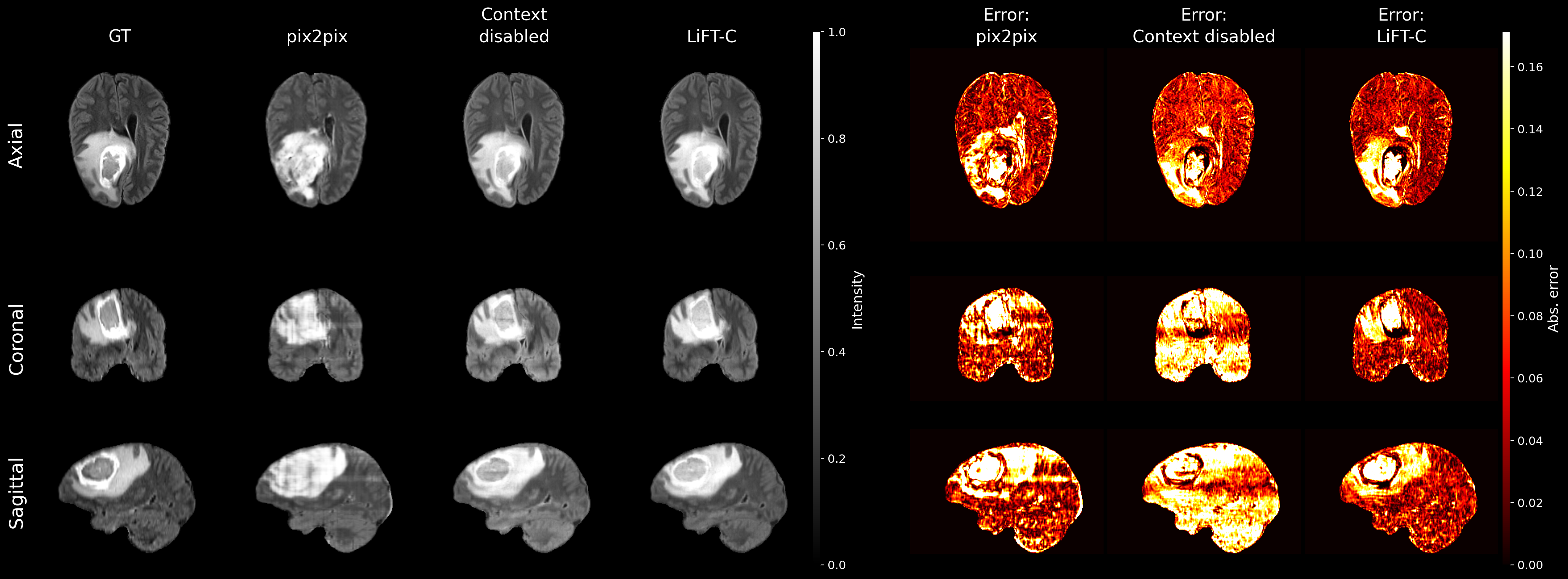}
\caption{Qualitative missing-MR comparison for a T2-FLAIR target.}
\label{fig:appendix_qual_missing_mr}
\end{figure}

Figure~\ref{fig:appendix_qual_missing_mr} provides a qualitative
comparison for missing-MR synthesis, including context-disabled
LiFT-C and corresponding error maps.

\section{Per-contrast missing-MR results}
\label{sec:appendix:missing_mr_percontrast}

Each target contrast is evaluated on the same $219$ subjects; the
\textit{Avg.} row is the unweighted mean across the four contrasts.

\begin{table}[!ht]
\centering
\small
\setlength{\tabcolsep}{3pt}
\caption{Per-contrast missing-modality MR synthesis on BraTS~2023 GLI.}
\label{tab:taskb_percontrast}
\begin{tabular}{lcc cc cc cc}
\toprule
& \multicolumn{2}{c}{pix2pix}
& \multicolumn{2}{c}{Fast-cWDM}
& \multicolumn{2}{c}{cWDM}
& \multicolumn{2}{c}{LiFT-C} \\
\cmidrule(lr){2-3}
\cmidrule(lr){4-5}
\cmidrule(lr){6-7}
\cmidrule(lr){8-9}
Target
& PSNR & SSIM
& PSNR & SSIM
& PSNR & SSIM
& PSNR & SSIM \\
\midrule
T1n & 27.607 & 0.9475 & 29.228 & 0.9568
    & 29.741 & 0.9622 & 29.497 & 0.9620 \\
T1c & 25.912 & 0.9316 & 26.951 & 0.9312
    & 27.316 & 0.9451 & 27.507 & 0.9482 \\
T2w & 26.694 & 0.9419 & 28.775 & 0.9560
    & 28.810 & 0.9588 & 28.643 & 0.9580 \\
T2f & 25.190 & 0.9169 & 27.980 & 0.9385
    & 27.831 & 0.9438 & 28.021 & 0.9430 \\
\midrule
Avg. & 26.351 & 0.9345 & 28.233 & 0.9456
     & 28.425 & 0.9525 & 28.417 & 0.9528 \\
\bottomrule
\end{tabular}
\end{table}

\subsection{Through-plane context ablation}
\label{sec:appendix:taskb_context}

We compare the same trained LiFT-C checkpoint with the context mixer
enabled or disabled at inference on the same $219$ subjects.
Table~\ref{tab:missing_mr_context} reports the unweighted mean across
the four target contrasts; lower values indicate better consistency
for all three metrics. Enabling context yields mean relative reductions
across contrasts of $48.4\%$, $3.9\%$, and $3.1\%$ in the axial-bias
SSIM gap, normalized adjacent-slice error, and absolute adjacent-slice
error, respectively.

\begin{table}[!ht]
\centering
\small
\caption{Through-plane context ablation for missing-MR synthesis.}
\label{tab:missing_mr_context}
\begin{tabular}{lccc}
\toprule
& Axial-bias SSIM gap
& Norm. adj.-slice error
& Abs. adj.-slice error \\
\midrule
Context disabled
& 0.00298
& 0.61672
& 0.00401 \\
LiFT-C
& 0.00144
& 0.59152
& 0.00388 \\
\bottomrule
\end{tabular}
\end{table}

Table~\ref{tab:taskb_context_stats} reports the per-contrast
axial-bias SSIM gap, absolute adjacent-slice error ($E_z$),
and normalized adjacent-slice error ($nE_z$) as mean$\pm$SD,
with reduction denoting relative improvement from
context-disabled inference to LiFT-C. All $12$ contrast-by-metric
paired comparisons remain significant after Bonferroni correction
($p_{\mathrm{adj}}\leq 1.88\times10^{-23}$). The largest relative
reduction in the axial-bias SSIM gap occurs for T2f ($74.7\%$).
The number of subjects improving with context ranges from
$184/219$ to $218/219$ across the $12$ comparisons.

\begin{table}[!ht]
\centering
\small
\setlength{\tabcolsep}{3pt}
\caption{Per-contrast Task-B context ablation.}
\label{tab:taskb_context_stats}
\begin{tabular}{llcccc}
\toprule
Target & Metric & Context off & Context on & Reduction & $p_{\mathrm{adj}}$ \\
\midrule
T1n & SSIM gap
& $0.00250\pm0.00317$ & $0.00192\pm0.00293$
& $23.3\%$ & $2.03{\times}10^{-36}$ \\
& n$E_z$
& $0.54854\pm0.11863$ & $0.54024\pm0.11881$
& $1.5\%$ & $4.03{\times}10^{-24}$ \\
& $E_z$
& $0.00392\pm0.00131$ & $0.00386\pm0.00133$
& $1.3\%$ & $1.88{\times}10^{-23}$ \\
\midrule
T1c & SSIM gap
& $0.00350\pm0.00333$ & $0.00145\pm0.00318$
& $58.6\%$ & $1.36{\times}10^{-36}$ \\
& n$E_z$
& $0.60454\pm0.07864$ & $0.58857\pm0.08763$
& $2.6\%$ & $5.82{\times}10^{-24}$ \\
& $E_z$
& $0.00396\pm0.00115$ & $0.00387\pm0.00120$
& $2.2\%$ & $1.11{\times}10^{-23}$ \\
\midrule
T2w & SSIM gap
& $0.00232\pm0.00316$ & $0.00146\pm0.00312$
& $36.9\%$ & $1.66{\times}10^{-35}$ \\
& n$E_z$
& $0.60332\pm0.08361$ & $0.58196\pm0.08875$
& $3.5\%$ & $1.77{\times}10^{-35}$ \\
& $E_z$
& $0.00397\pm0.00153$ & $0.00385\pm0.00156$
& $3.1\%$ & $5.25{\times}10^{-34}$ \\
\midrule
T2f & SSIM gap
& $0.00362\pm0.00309$ & $0.00091\pm0.00250$
& $74.7\%$ & $1.63{\times}10^{-36}$ \\
& n$E_z$
& $0.71050\pm0.17288$ & $0.65530\pm0.13164$
& $7.8\%$ & $6.21{\times}10^{-32}$ \\
& $E_z$
& $0.00418\pm0.00138$ & $0.00394\pm0.00146$
& $5.8\%$ & $3.91{\times}10^{-32}$ \\
\bottomrule
\end{tabular}
\end{table}

\section{Metric definitions}
\label{sec:appendix:metrics}

\paragraph{Standard image-quality metrics.}
Tasks~B and~C use the task-specific implementations of MAE, PSNR, SSIM,
and NCC described in the corresponding experiments. For Task~A, FID is
computed from $2048$-dimensional MedicalNet ResNet-50 features~\citep{medicalnet}
using $1{,}000$ generated and $1{,}000$ reference volumes following the
WDM protocol~\citep{Friedrich_2024}.

\paragraph{Through-plane metrics.}
Let $\Delta_z$ denote the forward adjacent-slice difference,
$(\Delta_z V)_{i,j,k}=V_{i,j,k+1}-V_{i,j,k}$. Task~C reports
$\Delta_z$ MAE between predicted and reference derivative volumes and
Pearson correlation between their vectorized derivatives. Region-restricted
$\Delta_z$ MAE is computed within the corresponding reference-CT mask.

\paragraph{Task-B reformat metric.}
Reformat SSIM is computed slice-wise in each orthogonal plane using 2D
SSIM~\citep{ssim} (window $11$, $\sigma=1.5$) and averaged within each
plane.

\paragraph{Paired statistical tests.}
For Task~B, context-enabled and context-disabled inference from the same
trained checkpoint are compared per subject using two-sided paired Wilcoxon
signed-rank tests, with Bonferroni correction over the $12$
contrast-by-metric comparisons. For Task~C, the context-disabled 2D translator
and LiFT-C are compared on the same $36$ subjects using
two-sided exact Wilcoxon signed-rank tests, with Bonferroni correction over
the eight reported conventional and through-plane metrics. For the all-model
Task-C comparison, LiFT-C is additionally compared with each of the three
baselines using two-sided paired Wilcoxon signed-rank tests on the same $36$
subjects, with Bonferroni correction over the $12$ baseline-by-metric
comparisons.

\paragraph{CT region masks.}
Task~C masks are derived from the reference CT: bone is
$\mathrm{HU}>300$, air is $\mathrm{HU}<-500$, and the body mask is the
largest connected component of $\mathrm{HU}>-500$ with internal holes
filled~\citep{yang2025hu,lee2025abdominal}. CT volumes are clipped to
$[-1000,2000]$\,HU before evaluation.

\section{Paired MR-to-CT ablation statistics}
\label{sec:appendix:mrct_stats}

We compare LiFT-C with its context-disabled 2D translator on the same
$36$ SynthRAD2023 model-selection/evaluation subjects. Conventional
metrics use the official SynthRAD masked evaluator, whereas the
$\Delta_z$ metrics use our adjacent-slice coherence evaluator. Positive
improvement below always denotes better LiFT-C performance.

All $36$ subjects improve on all eight reported metrics. Two-sided exact
Wilcoxon signed-rank tests remain significant after Bonferroni correction
over the eight comparisons
($p_{\mathrm{adj}}=2.33\times10^{-10}$ for each). The identical adjusted
$p$-values reflect the resolution of the exact test at $N=36$, rather than
equal effect sizes across metrics.

\begin{table}[!ht]
\centering
\small
\setlength{\tabcolsep}{3pt}
\caption{Paired MR-to-CT context ablation.
Positive values indicate improvement over context-disabled inference.}
\label{tab:taskc_paired_stats}
\begin{tabular*}{\linewidth}{@{\extracolsep{\fill}}l*{8}{c}@{}}
\toprule
& \multicolumn{4}{c}{Image quality}
& \multicolumn{4}{c}{Through-plane coherence} \\
\cmidrule(lr){2-5}
\cmidrule(lr){6-9}
Improvement
& \shortstack{MAE\\(HU)}
& \shortstack{PSNR\\(dB)}
& SSIM
& NCC
& \shortstack{$\Delta_z$ MAE\\Full (HU)}
& \shortstack{$\Delta_z$ MAE\\Bone (HU)}
& \shortstack{$\Delta_z$ MAE\\Air (HU)}
& \shortstack{$\Delta_z$\\correlation} \\
\midrule
Absolute
& 1.914 & 0.255 & 0.0054 & 0.0078
& 4.364 & 16.760 & 22.491 & 0.0534 \\
Relative (\%)
& 3.42 & 0.88 & 0.58 & 0.89
& 10.21 & 11.05 & 13.06 & 7.91 \\
\bottomrule
\end{tabular*}
\end{table}

\section{Memorization probe}
\label{sec:appendix:memorization}

For unconditional generation, we assess near-exact reproduction of
training volumes using nearest-neighbor retrieval. For each of $200$
LiFT-U-generated volumes, we compute the mean voxel-wise L1 distance
to its nearest training volume in normalized MR space. The
nearest-neighbor distance is $0.063\pm0.007$ for generated volumes,
compared with $0.026\pm0.004$ for the held-out $219$-subject cohort
(mean$\pm$SD). These results provide no evidence of near-exact
training-volume copying among the evaluated samples.

\section{Datasets and licenses}
\label{sec:appendix:licenses}

All datasets and pretrained weights used in this work are publicly released.

\paragraph{BraTS 2023 GLI.}
Tasks~A and~B use the de-identified, skull-stripped BraTS~2023 GLI cohort
distributed through the RSNA-ASNR-MICCAI challenge under its academic,
non-commercial data-use agreement.

\paragraph{SynthRAD2023 Task 1 -- Brain.}
Task~C uses the publicly released, co-registered and de-identified
SynthRAD2023 MR--CT pairs~\citep{synthrad2023} under its public
non-commercial research terms. We use a fixed
patient-level split of $144$ training and $36$ model-selection/evaluation
subjects.

\paragraph{Pretrained models and baselines.}
The MedicalNet ResNet-50~\citep{medicalnet} weights, released under the MIT
license, are used only as a frozen feature extractor for Task-A FID. pix2pix~\citep{pix2pix} is reproduced
from the official implementation, and cWDM/Fast-cWDM~\citep{cwdm} use the
official implementation on the same $219$-subject Task-B cohort; Fast-cWDM
uses $100$ sampling steps and cWDM uses $1000$.

\section{Training hyperparameters}
\label{sec:appendix:hyperparams}

Table~\ref{tab:train_hparams} summarizes the optimizer, learning rate,
and batch construction for each training stage.

\begin{table}[!ht]
\centering
\small
\setlength{\tabcolsep}{4pt}
\renewcommand{\arraystretch}{1.15}
\caption{Training settings for LiFT experiments.
For missing-MR synthesis, learning rates are listed as
translator/context module.}
\label{tab:train_hparams}
\begin{tabularx}{\linewidth}{@{}
    >{\raggedright\arraybackslash}X
    >{\raggedright\arraybackslash}X
    >{\raggedright\arraybackslash}X
    >{\raggedright\arraybackslash}X
@{}}
\toprule
Task / stage & Trainable module & Optimizer / LR & Batch construction \\
\midrule
LiFT-U: slice generator
& 2D axial generator
& Adam\newline $2{\times}10^{-4}$
& $128$ axial slices \\

LiFT-U: depth mapper
& Depth mapper
& Adam\newline $2{\times}10^{-4}$
& $8$ volumes \\

Missing-MR: LiFT-C
& 2D U-Net + spatial BiGRU
& AdamW\newline $2{\times}10^{-5}$ / $2{\times}10^{-4}$
& $1$ volume; $48$ decoded slices \\

MR-to-CT: translator
& 2D U-Net
& Adam\newline $2{\times}10^{-4}$
& $32$ axial slices \\

MR-to-CT: context refinement
& Spatial BiGRU + residual head
& AdamW\newline $2{\times}10^{-4}$
& $32$ three-slice windows \\
\bottomrule
\end{tabularx}
\end{table}

For LiFT-U, tri-planar drift uses a frozen ImageNet-pretrained
ResNet-18, with real reference features drawn from the current
minibatch. Inference uses sampling temperature $T=1.25$.

Missing-MR synthesis uses a weighted combination of L1, SSIM,
VGG perceptual, and depth-consistency losses. Reconstruction and SSIM losses use all decoded slices, VGG loss
uses $8$ sampled slices, and the depth loss uses successive pairs
in the sampled subset.

For MR-to-CT synthesis, the 2D translator is frozen during context
refinement. Context spans all $128$ slices; the three-slice windows
in Table~\ref{tab:train_hparams} specify the decoded supervision subset.

Tasks~A and~B use BraTS validation criteria for checkpoint selection;
Task~C uses the fixed $36$-subject model-selection/evaluation cohort.

\section{Algorithms}
\label{sec:appendix:algorithms}
\begin{algorithm}[!ht]
\caption{LiFT-U training step}
\label{alg:liftu_training}
\begin{algorithmic}[1]
\Require Latents $\{\xi_i\}_{i=1}^{B}$,
real volumes $\{V_i\}_{i=1}^{B}$,
frozen slice generator $G_{\mathrm{2D}}$,
depth mapper $M_\phi$,
tri-planar objective $\mathcal{L}_{\mathrm{LiFT-U}}$
\Ensure Updated depth-mapper parameters $\phi$

\For{$i=1,\ldots,B$}
    \For{$d=1,\ldots,D$}
        \State $c_{i,d} \gets M_\phi(\xi_i,\gamma(d))$
        \State $\hat v_{i,d} \gets G_{\mathrm{2D}}(c_{i,d})$
    \EndFor
    \State $\hat V_i \gets
        \operatorname{Stack}_{d=1}^{D}\hat v_{i,d}$
\EndFor

\State $\mathcal{L} \gets
    \mathcal{L}_{\mathrm{LiFT-U}}
    \bigl(\{\hat V_i\}_{i=1}^{B},\{V_i\}_{i=1}^{B}\bigr)$
\State Update $\phi$ using $\nabla_\phi\mathcal{L}$

\end{algorithmic}
\end{algorithm}

\begin{algorithm}[!ht]
\caption{LiFT-C training step}
\label{alg:liftc_training}
\begin{algorithmic}[1]
\Require Source--target pairs $\{(X_i,Y_i)\}_{i=1}^{B}$,
2D encoder $E_\theta$, 2D decoder $D_\theta$,
inter-slice context module $M_\phi^C$,
residual mapping $R_\phi$,
supervised objective $\mathcal{L}_{\mathrm{sup}}$,
trainable parameters $\Theta_{\mathrm{train}}\subseteq\{\theta,\phi\}$
\Ensure Updated trainable parameters $\Theta_{\mathrm{train}}$

\For{$i=1,\ldots,B$}
    \For{$d=1,\ldots,D$}
        \State $h_{i,d} \gets E_\theta(X_{i,d})$
    \EndFor
    \State $(c_{i,1},\ldots,c_{i,D}) \gets
        M_\phi^C(h_{i,1:D})$
    \For{$d=1,\ldots,D$}
        \State $\tilde h_{i,d} \gets h_{i,d}+R_\phi(c_{i,d})$
        \State $\hat y_{i,d} \gets D_\theta(\tilde h_{i,d})$
    \EndFor
    \State $\hat Y_i \gets
        \operatorname{Stack}_{d=1}^{D}\hat y_{i,d}$
\EndFor

\State $\mathcal{L} \gets
    \mathcal{L}_{\mathrm{LiFT-C}}
    \bigl(\{\hat Y_i\}_{i=1}^{B},\{Y_i\}_{i=1}^{B}\bigr)$
\State Update $\Theta_{\mathrm{train}}$ using
    $\nabla_{\Theta_{\mathrm{train}}}\mathcal{L}$

\end{algorithmic}
\end{algorithm}

% \clearpage
% \input{checklist.tex}
\end{document}